\documentclass[runningheads]{llncs}

\usepackage{eccv}

\usepackage{eccvabbrv}

\usepackage{graphicx}
\usepackage{booktabs}

\usepackage[accsupp]{axessibility}  % Improves PDF readability for those with disabilities.

\usepackage{inconsolata}
\usepackage{kotex}
\usepackage{bm}
\usepackage{booktabs}
\usepackage[export]{adjustbox}
\usepackage{hhline}
\usepackage{multirow}
\usepackage{comment}

\usepackage{subcaption}
\usepackage{tabularx}

\usepackage{hyperref}

\usepackage[capitalize]{cleveref}
\crefname{section}{Sec.}{Secs.}
\Crefname{section}{Section}{Sections}
\Crefname{table}{Table}{Tables}
\crefname{table}{Tab.}{Tabs.}

\definecolor{azure}{rgb}{0.0, 0.5, 1.0}
\definecolor{awesome}{rgb}{1.0, 0.13, 0.32}
\definecolor{forestgreen}{rgb}{0.13, 0.55, 0.13}

\usepackage{orcidlink}

\begin{document}

%\title{Contexts through Structural Insights:\\
%Rethinking the Roles of Geometric Information\\in 3D Dense Captioning}
% \title{Role-Specific Query Segregation and\\Contextualization from Geometry\\for 3D Dense Captioning}
\title{Bi-directional Contextual Attention for\\3D Dense Captioning}

\titlerunning{Bi-directional Contextual Attention for 3D Dense Captioning}
\newcommand{\modelname}{BiCA}

\makeatletter
\renewcommand{\thefootnote}{\fnsymbol{footnote}}
\def\@fnsymbol#1{\ensuremath{\ifcase#1\or *\or \dagger\or \ddagger\or
   \mathsection\or \mathparagraph\or \|\or **\or \dagger\dagger
   \or \ddagger\ddagger \else\@ctrerr\fi}}
\makeatother
\newcommand*\samethanks[1][\value{footnote}]{\footnotemark[#1]}

\author{Minjung Kim\inst{1,4}$^ ,$\thanks{Work done during internship at LG AI Research}\orcidlink{0009-0009-8038-3710} \and
Hyung Suk Lim\inst{3}\orcidlink{0009-0006-0810-1027} \and
Soonyoung Lee\inst{2} \and \\
Bumsoo Kim\inst{2}$^ ,$\thanks{Corresponding authors} \and
Gunhee Kim\inst{1}$^ ,$\samethanks
}

\authorrunning{M. Kim et al.}

\institute{Seoul National University, Seoul, Korea \\ \email{minjung.kim@vision.snu.ac.kr, gunhee@snu.ac.kr} \\
\and LG AI Research, Seoul, Korea \\ \email{\{soonyoung.lee,bumsoo.kim\}@lgresearch.ai} \\
\and Diquest, Seoul, Korea \\ \email{hslim@diquest.com} \\
\and SNU-LG AI Research Center, Seoul, Korea}

\maketitle

\begin{abstract}
  3D dense captioning is a task involving the localization of objects and the generation of descriptions for each object in a 3D scene. Recent approaches have attempted to incorporate contextual information by modeling relationships with object pairs or aggregating the nearest neighbor features of an object. However, the contextual information constructed in these scenarios is limited in two aspects: first, objects have multiple positional relationships that exist across the entire global scene, not only near the object itself. Second, it faces with contradicting objectives--where localization and attribute descriptions are generated better with tight localization, while descriptions involving global positional relations are generated better with contextualized features of the global scene. To overcome this challenge, we introduce {\modelname}, a transformer encoder-decoder pipeline that engages in 3D dense captioning for each object with Bi-directional Contextual Attention. Leveraging parallelly decoded instance queries for objects and context queries for non-object contexts, BiCA generates object-aware contexts, where the contexts relevant to each object is summarized, and context-aware objects, where the objects relevant to the summarized object-aware contexts are aggregated. This extension relieves previous methods from the contradicting objectives, enhancing both localization performance and enabling the aggregation of contextual features throughout the global scene; thus improving caption generation performance simultaneously. Extensive experiments on two of the most widely-used 3D dense captioning datasets demonstrate that our proposed method achieves a significant improvement over prior methods.
\end{abstract}
\section{Introduction}
\label{sec:intro}

%3D dense captioning is a task where we need to 1) localize all objects and 2) generate descriptive sentences for each detected object within a 3D scene.
3D dense captioning is a task that requires 1) determining the location of all objects and 2) generating descriptive sentences for each object detected within a 3D scene.
Early approaches have utilized a two-stage process~\cite{chen2021scan2cap, jiao2022more, wang2022spacap3d, zhong2022contextual, cai20223djcg, chen2022d3net, yuan2022xtrans2cap}, where the objects are detected first, then captions are generated afterward by sequentially feeding the features of each detected object.
Following work~\cite{chen2023vote2capdetr} has borrowed the end-to-end transformer encoder-decoder pipeline from object detection~\cite{carion2020detr} to actively involve contextual information for 3D dense captioning.
In these works, contextual information for predicting relationships is composed by combining object features (e.g., with dot product~\cite{wang2022spacap3d,cai20223djcg} or transformer attention~\cite{chen2023vote2capdetr}) and aggregating local nearest neighbor visual features~\cite{chen2023vote2capdetr}.

Despite their improved performance, the scope of contextual information in previous 3D dense captioning methods poses its own set of challenges.
First, 3D dense captioning heavily involves various positional relationships throughout the global scene; thus, the context scope restricted to the combination of individual objects or their spatially nearest neighbors typically cannot provide sufficient information.
Next, the endeavor to incorporate the contextual features of various positions into a single object encounters a conflicting objective since the object must encompass precise local features for localization while simultaneously capturing various positional relationships throughout the global context for caption generation.
Clearly, these challenges limit the performance of 3D dense captioning, which raises an interesting research question-- \textit{``Can we design an architecture for 3D dense captioning that can effectively aggregate relevant context features without harming localization performance?"}

\begin{figure*}[t]
\begin{center}
\centerline{\includegraphics[width=1.0\textwidth]{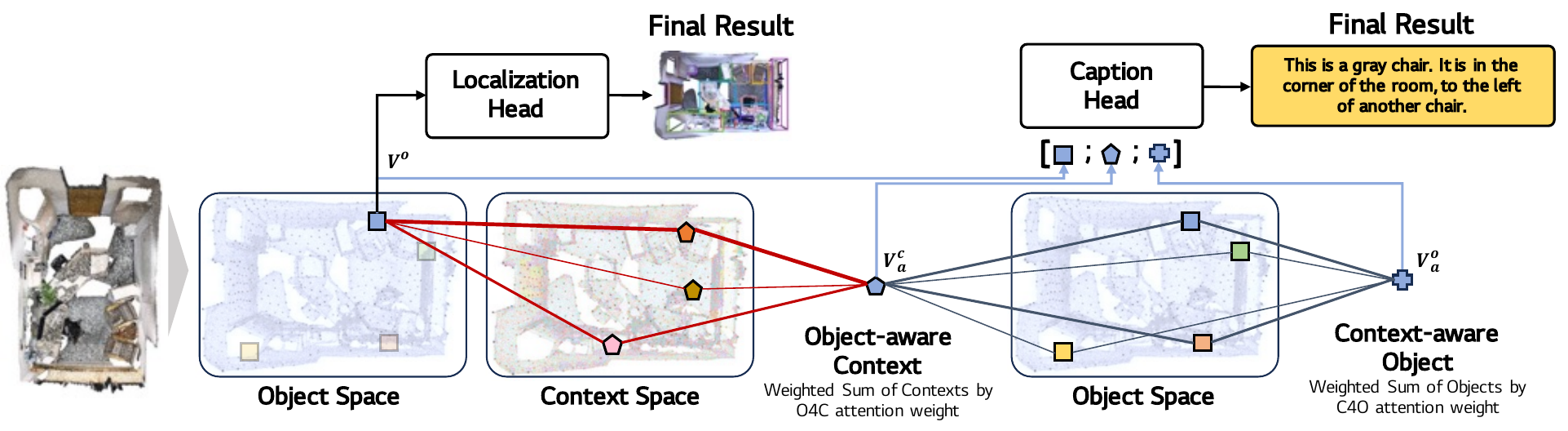}}
\caption{
Conceptual illustration of the multi-stage pipeline of {\modelname} (best viewed in color). %engaging contextualized geometry and contextualized objects.
}
\label{fig:concept}
\end{center}
\end{figure*}
To this end, we propose a \textbf{Bi}-directional \textbf{C}ontextual \textbf{A}ttention (BiCA) network for 3D dense captioning.
BiCA parallelly decodes a fixed set of \textit{object} queries and \textit{context} queries sampled from the geometric positions that do not overlap with each other.
While the decoded object query performs standard object detection, the context queries are designed to distinctively capture \textit{non-object} contexts and subsequently aggregate relevant contextual information for each object.
Recognizing that non-object regions often lack sufficient visual features in 3D point clouds, BiCA incorporates a multi-stage pipeline where context features are enriched by relevant object features, thus implementing \textit{bi-directional} attention: i) Contextual Attention of Objects for Context (O4C): context representation is represented by the attention summarization of global objects. ii) Contextual Attention of Contexts for Object (C4O): the object representation is adjusted by summarizing the non-object contexts.

Then, localization is inferred with object queries while the captions are generated with object-aware context and context-aware object features.
By disentangling the attention for localization and the attention for incorporating contextual features, BiCA successfully resolves the aforementioned contradicting objectives for local and global features.
Moreover, since the contextual attentions retrieve globally relevant contexts and objects throughout the scene, it could accurately produce descriptions that transcend the spatial boundary of localization and its nearest neighbor.
The conceptual overview of our BiCA is illustrated in \Cref{fig:concept}.

{\modelname} sets a new state-of-the-art standard for 3D dense captioning, while simultaneously improving 3D object detection performance.
Extensive experiments on two widely used benchmarks in 3D dense captioning (i.e., ScanRefer~\cite{chen2020scanrefer} and Nr3D~\cite{achlioptas2020referit3d}) show that our proposed {\modelname} surpasses prior approaches by a large margin.
The contribution of our paper can be summarized as:
\begin{itemize}
    \item We propose a novel Bi-directional Contextual Attention network with multi-stage contextual attention for 3D dense captioning. 
    This enables our model to capture relevant contexts throughout the global scene without being bound to single-object localization or their nearest neighbors. 
    \item By performing localization with decoded object queries and caption generation by including object-aware context features and context-aware object correspondence features, {\modelname} can simultaneously improve the performance of localization and caption generation for 3D dense captioning.
    \item Our {\modelname} achieves state-of-the-art performances across multiple evaluation metrics on two widely used benchmarks for 3D dense captioning: ScanRefer and Nr3D datasets.
\end{itemize}

\section{Related work}
\label{sec:related_work}

\begin{comment}
In this section, we summarize a comprehensive overview of representative studies on image captioning and vision-language tasks closely related to 3D dense captioning research. We then explore various techniques developed for 3d dense captioning, highlighting their methodologies and contributions and introducing methods for understanding and describing complex 3D environments.
\end{comment}

\subsection{Image Captioning}
Image captioning is a fundamental task in visual language creation that automatically generates descriptive sentences for images. 
The main goal is to improve 2D scene understanding and generate captions that accurately reflect the content and context of the image~\cite{hossain2019comprehensive}.
%In the early stages of image captioning, the dominant techniques centered around template-based approaches \cite{farhadi2010every} that formed sentences based on predefined structures, and retrieval-based mechanisms \cite{wu2017automatic}, where the most suitable sentence was selected from a repository of potential captions for a specific image.
However, with the advancement of deep learning in image captioning, there has been a marked shift towards the development and adoption of more sophisticated models~\cite{ghandi2022deep}.
Image captioning methods using deep learning adopt an encoder-decoder architecture, where the decoder generates a sentence from the visual features extracted by the encoder. 
Attention-based methods for grid regions~\cite{xu2015show} and detected objects~\cite{anderson2018bottom} focus on specific image regions and use graph neural networks~\cite{yang2019auto} or transformer layers to capture relationships between objects~\cite{cornia2020meshed}.
\subsection{3D Dense Captioning}

3D dense captioning, an emerging field focused on achieving detailed object-level understanding of 3D scenes through natural language descriptions, has garnered significant interest in recent years.
This task involves converting 3D visual data~\cite{dai2017scannet} into a consistent set of bounding boxes and generating appropriate natural language descriptions for each identified object. %presenting a considerable challenge due to the complexity of 3D scene information.
This task presents a considerable challenge due to the complexity of 3D scene information.

Most 3D dense captioning models employ an encoder-decoder architecture comprising three main components: a scene encoder, a relational module, and a feature decoder.
The scene encoder in 3D dense captioning utilizes 3D object detection methods such as 3DETR~\cite{3detr}, PointNet++~\cite{qi2017pointnet++}, VoteNet~\cite{votenet}, and PointGroup~\cite{jiang2020pointgroup} to extract object-level visual features and contextual details from 3D point cloud datasets.
%VoteNet has been the preferred backbone for feature extraction in early 3D dense captioning methods \cite{chen2021scan2cap,jiao2022more,wang2022spacap3d,yuan2022xtrans2cap,cai20223djcg}. Recent advancements, such as the incorporation of PointGroup in D3Net \cite{chen2022d3net} and the adoption of an end-to-end Transformer in Vote2Cap-DETR \cite{chen2023vote2capdetr}, have further improved the accuracy and contextual understanding in 3D scene interpretation and captioning.
The relational module is crucial in 3D dense captioning, modeling complex connections and cross-modal interactions among objects within indoor scenes. Relational modeling approaches vary based on task requirements and can be categorized into graph-based~\cite{chen2021scan2cap, jiao2022more, chen2022d3net}, transformer-based~\cite{cai20223djcg, wang2022spacap3d, chen2023vote2capdetr}, and knowledge distillation-based~\cite{yuan2022xtrans2cap} methods.
The feature decoder generates bounding boxes and captions for candidate objects by considering attributes and relational features. 
GRU-based decoders with attention mechanisms are used in Scan2Cap~\cite{chen2021scan2cap}, MORE~\cite{jiao2022more}, and D3Net~\cite{chen2022d3net}, 
while transformer-based decoders are employed in SpaCap3D~\cite{wang2022spacap3d}, $\chi$-Tran2Cap~\cite{yuan2022xtrans2cap}, 3DJCG~\cite{cai20223djcg}, and Vote2Cap-DETR~\cite{chen2023vote2capdetr} to facilitate the caption generation process.

%However, most existing methods have faced several challenges as they used a single set of queries for object localization and caption generation, often leading to conflicts in effectively achieving both tasks.
However, most existing methods use a single query set for object localization and caption generation, which often leads to conflicts in achieving the objectives of the two tasks.
To address this, Vote2Cap-DETR++~\cite{chen2024vote2cap} enhances task-specific feature capture by decoupling queries for localization and captioning, overcoming the limitations of a unified query system. 
Despite this improvement, challenges persist as its performance remains fundamentally bound by the precision of its object localization capabilities.

\section{Method}
\label{sec:method}
\begin{figure*}[t]
\begin{center}
\centerline{\includegraphics[width=1.0\textwidth]{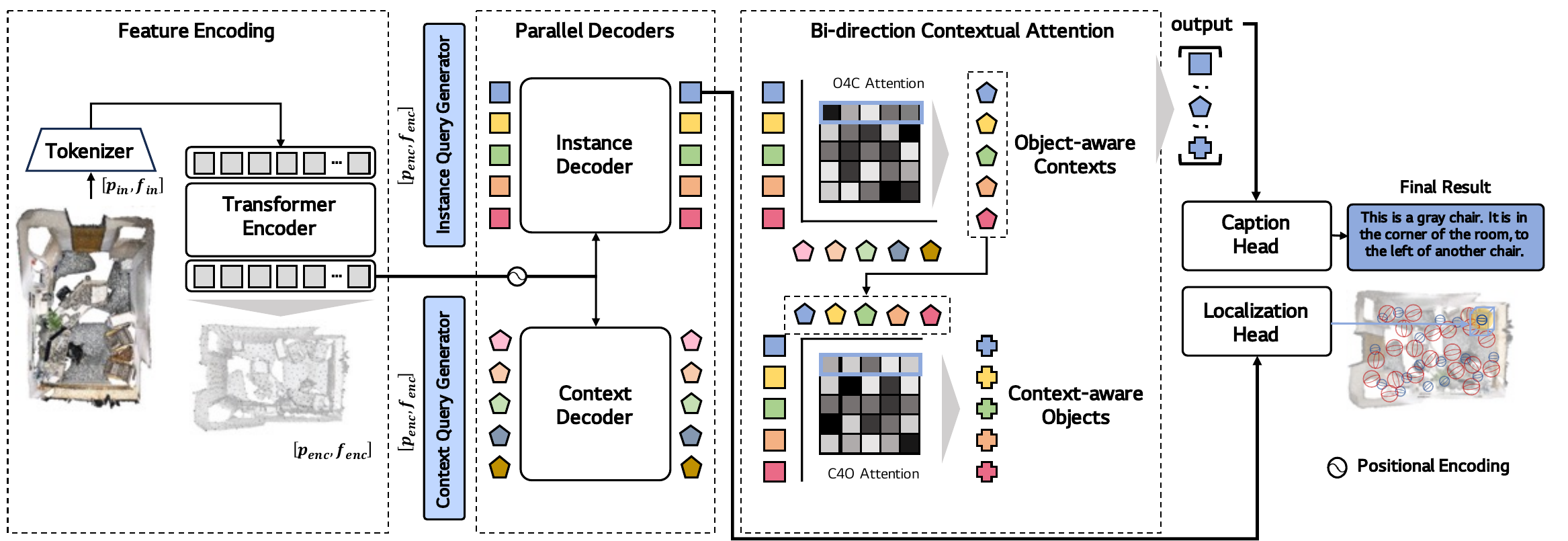}}
\caption{
The overall pipeline of {\modelname}. We parallelly generate and decode two sets of queries (i.e., Instance Query and Context Query) that encodes the instance features and the non-object context features throughout the global scene, respectively. The object-aware contexts are calculated per each object by the weighted sum of the context queries, where the weights are calculated by the attention between the decoded instance query and context query. Then, with the object-aware context feature, the context-aware object feature is obtained by the weighted sum of the instances, which is weighted by the attention between the object-aware contexts.
}
\label{fig:overall}
\end{center}
\end{figure*}

Our goal is to take a 3D scene as input, identify all the objects within it, and generate captions for each object.
%In previous research, a model is learned to identify objects in a 3D scene and simultaneously generate captions from the same features.
Previous research has adopted a pipeline to identify objects in a 3D scene and simultaneously generate captions from the same features.
%Each caption may describe the characteristics of the object itself, but may also be written based on relationships with surrounding objects or information about the entire scene.
However, while each caption may describe features of the object itself, it can also be based on the relationships with surrounding objects or information about the overall scene.

This section introduces a novel transformer encoder-decoder approach with bi-directional contextual attention for the 3D dense captioning task.
This approach distinguishes between features for instances and context information between instances, indirectly supervising the context features to enable end-to-end learning with the instance features.
Instance queries aim to localize instances within a 3D scene and capture the characteristics of instances.
In contrast, each context query is designed to capture the non-object region within a scene, including positional relationships between instances in the 3D scene.
The overall architecture of the newly proposed model is described in \Cref{fig:overall}.

\subsection{Encoder}

Following previous work~\cite{chen2023vote2capdetr}, our model also adopts 3DETR~\cite{3detr} encoder as our scene encoder.
This encoder is applied to transform the input point cloud into a set of encoded tokens that capture spatial, structural, and contextual information.
Given the input point cloud \(PC = [p_{\text{in}}; f_{\text{in}}] \in \mathbb{R}^{N \times (3 + F)}\), it is initially tokenized by a set-abstraction layer of PointNet++~\cite{qi2017pointnet++}.
This tokenized output is subsequently fed into a masked transformer encoder incorporating the set-abstraction layer, followed by two additional encoder layers.
The final encoded scene tokens are denoted as \(p_{enc} \in \mathbb{R}^{1,024 \times 3}\) and \(f_{enc} \in \mathbb{R}^{1,024 \times 256}\).
% Afterwards, our proposed decoder system with region-centric attention decomposition parallely decodes two sets of query sets that each captures individual object regions and regions that cover the contextual relationship between multiple objects.
\subsection{Query Generator}

To separate captions bound to a single object from captions containing information relative to other objects or the global scene, we specify two separate \textit{Instance Query} and \textit{Context Query} from the encoded scene.
%The instance query is decoded to perform standard object localization, while the anchor query is decoded to generate features for unique geometries in the scene.
While the instance query set is designed to capture individual features of objects bounded by their localization, 
the context query set is designed to encompass a broad area that consists of the \textit{positional information} of a caption (e.g., next to, on the right of, is most far from, etc.).
% While the context query captures the contextual regions capable of captioning, the \textit{instance query} generates standard object localization and attribute-related caption prediction for each object.
% The two queries are decoded in parallel. %and latter aggregated to consist the final caption.

\subsubsection{Instance Query Generator.}
%As the context query captures the contextual information for each caption, 
%The instance query is extracted to perform standard 3D object detection and generate captions for the individual attributes of each object.
In the Instance Query Generator, instance queries are generated to detect all objects in the 3D scene and generate captions for each object's individual attributes.
The instance query $(p^o,f^o)$ is defined as follows:
\begin{equation}
    [\Delta p_{\text{vote}}; \Delta f_\text{vote}]=\text{FFN}_{o}(f_{\text{enc}}),
\end{equation}
\begin{equation}
    (p^o,f^o)=\text{SA}_o\big(p_{\text{enc}}+\Delta p_{\text{vote}}, f_\text{enc}+\Delta f_\text{vote}\big),
\end{equation}
where $[\Delta p_{\text{vote}}; \Delta f_\text{vote}]\in\mathbb{R}^{1,024\times(3+256)}$ is an offset that learns to shift the encoded points to object's centers spatially by a feed-forward network $\text{FFN}_{o}$, following \cite{chen2023vote2capdetr}.
$\text{SA}_o$ represents the set-abstraction layer with a radius of $0.3$ and samples $16$ points for $p^o$.
Since our Instance Query Generator extracts the features from the candidate coordinates after the voting, our instance queries are not focused on specific objects. 
However, they are distributed across the locations where objects are present throughout the 3D scene. 
As a result, $256$ instance queries are extracted from our Instance Query Generator.

\subsubsection{Context Query Generator.}
3D scenes are composed of 3D points arranged in XYZ coordinates, forming specific objects, and the spaces between objects signify space.
We recognize that obtaining context features should not merely involve interpolating the spaces between Instance Queries but also encompassing how these points are arranged and the composition of empty spaces to represent a scene. 
Inspired by this observation, we define \textit{context information} as the geometric details in a 3D scene that illustrate the relationships between objects and between objects and the scene itself.
We decide to extract structural information centered on specific points in the 3D scene and call them \textit{Context Queries}.
This approach is distinct from the Instance Query Generator. 
While the Instance Query Generator uses a voting network to obtain object center coordinates that do not exist in the 3D scene and extracts features such as identification from them, the proposed Context Query Generator extracts structural information without disrupting the arrangement of the 3D points.

Given the encoded scene tokens $(p_{\text{enc}}, f_{\text{enc}})$, we sample $512$ points $p^c_{\text{seed}}$ with farthest point sampling (FPS)~\cite{qi2017pointnet++} on $p_{\text{enc}}$ that each represent a unique geometry within the global scene. 
The context query $(p^c,f^c)$ is then defined as:
\begin{equation}
    (p^c,f^c)=\text{SA}_c(p_{\text{enc}},f_{\text{enc}}),
\end{equation}
where $\text{SA}_c$ denotes the set-abstraction layer~\cite{qi2017pointnet++} with a radius of $1.2$, sampling $64$ points for $p^c$.
All hyper-parameters are determined experimentally.
\subsection{Decoder}

\subsubsection{Instance Decoder and Context Decoder.}
%Given the anchor queries $(p^c,f^c)$ and instance queries $(p^o,f^o)$, we build a parallel decoding pipeline where the context decoder designates contextual information between objects, and the localization and attribute description for the participating objects is performed by the instance decoder.
Given the context queries $(p^c,f^c)$ and instance queries $(p^o,f^o)$, we build a parallel decoding pipeline where the Context Decoder designates contextual information between objects and the Instance Decoder performs the localization and attribute description for the participating objects.

Each Instance Decoder and Context Decoder consists of a transformer decoder with $8$ layers according to 3DETR~\cite{3detr}, and Fourier positional encoding is applied to XYZ coordinates.
Fourier positional encoding transforms the XYZ coordinates of each query into a Fourier position embedding~\cite{tancik2020fourier} and then combines them with the corresponding query embeddings. 
This encoding transforms the input data into frequency functions, capturing invariant changes in complex shapes and structures. 
It aids the context queries in accurately identifying geometric shapes and the instance queries in detecting objects. 
The context queries $V^c$ are afterwards contextualized for each instance query $V^o$, then retrieve relevant instance query features that are structurally involved.
% We then feed the decoded context query $V^c$ and instance query $V^o$ to our Context-Aggregator \mj{check module name} so that we can learn the caption by combining the context feature with the surrounding instance features.
\subsection{Bi-directional Contextual Attention}

%In 3D dense captioning, context information does not only consist of relationships with surrounding objects but is also expressed as relationships with the entire space and relationships between the most distant objects.
Captions must be generated for each object in relation to the scene context, and through positional encoding, it becomes possible to understand directional relationships, such as "to the right" with surrounding objects. 
Here, we gain the insight that information about the relative objects closely related to these relationships is also necessary, and we design the model to proceed simultaneously with a multi-stage contextualization.
With the decoded queries, our method proceeds in two stages: i) Contextual Attention of Objects for Context (O4C): context representation is represented by the attention summarization of global objects. ii) Contextual Attention of Contexts for Object (C4O): the object representation is adjusted by summarizing the non-object contexts.
\Cref{fig:concept} and \Cref{fig:overall} show detailed illustrations of obtaining the object-aware context feature and the context-aware object feature.

\subsubsection{Contextual Attention of Objects for Context (O4C).}
To retrieve the structural relationship between objects throughout the global scene, we first construct a Object-aware Context $V^c_a$ per object by summating the attention of each context position feature with regard to each object.
The attention result between the instance feature and the context feature is applied to the context feature, making a weighted summarization of the geometries.
At this time, the ratio is adjusted using learnable gamma.
We name this result $V^c_a$.

\subsubsection{Contextual Attention of Contexts for Object (C4O).}
Most contexts not only interact with the reference object but also have other objects that react against them.
To achieve this, we go through a Contextual Attention of Contexts for Object process.
This could specify the information that can simply be said to be ‘next to’ as ‘next to the red chair’.
To this end, the attention result of the context feature and instance feature with the attention weight applied is applied to the instance query feature.
Likewise, it is adjusted using a learnable parameter lambda.
We name this result $V^o_a$.
$V^o_a$ results from finding instance combinations that have a meaningful relationship with the contextual region focused on a specific instance.
The concatenation of the instance token $V^o$, contextualized geometry $V^c_a$, and contextualized object $V^o_a$, denoted as $V^a$, is fed to the caption head to generate final captions.
\subsection{3D Dense Captioning}

Our final goal is to identify all objects within the input 3D scene and generate a caption for each object.
We perform object detection and caption generation in parallel and group the results into one instance.

\subsubsection{Localization.}
To localize instances in the 3D scene, we use the decoded instance query $V^o$. 
%and parallely feed it into the detection head and (shared) caption head.
Through $5$ MLP heads, we reformulate the box corner estimation as offset estimation from a query point to an object’s center and box size regression.
%All subtasks are implemented by FFNs.
These localization heads are shared across the decoder layers.

\subsubsection{Caption Generation.}
For caption generation, we use a transformer decoder-based caption head based on GPT-2~\cite{gpt2}, following the methods of Vote2Cap-DETR~\cite{chen2023vote2capdetr} and SpaCap3D~\cite{wang2022spacap3d}.
Our caption head comprises two transformer decoder blocks, sinusoid positional embedding, and a linear classification layer.
When generating captions for a proposal, we substitute the standard Start Of Sequence ('SOS') prefix with $V^a$.
During the inference process, our approach employs beam search to produce captions.
The beam size is $5$.
\subsection{Training {\modelname}}
%In this section, we include details of how to train \edit{our method}. %end-to-end transformer encoder-decoder pipeline with caption-centric attention, SIA (i.e., See-it-All).
%To compare with previous object-centric methods using benchmark datasets, we construct the final caption to be object-centric.
%The final caption for the $i$-th object is obtained by simply concatenating the captions generated from  $V^o_i$ and $V^a_i$.
%Our method is trained and evaluated by locating all objects within a scene and comparing the final caption centered on each object with the ground-truth.

\subsubsection{Instance Query Loss.}
%Identical to previous work~\cite{chen2023vote2capdetr}, we borrow the vote loss from VoteNet~\cite{votenet} to help the Instance Query Generator learn to shift points $p_{enc}$ to an object's center.
To train instance query generator to find an object's center by shifting points $p_{enc}$, we use the vote loss from VoteNet~\cite{votenet}.
Given the generated instance query $(p^o,f^o)$ and the encoded scene tokens $(p_{enc},f_{enc})$, the vote loss $\mathcal{L}^o$ is denoted as:
\begin{equation}
    \mathcal{L}^o=\frac{1}{M}\sum_{i=1}^M\sum_{j=1}^{N_{gt}}{\Vert p^o_i-{\text{cnt}}_j\Vert_1\cdot \mathbb{I}(p^i_{enc})},
\end{equation}
where $\mathbb{I}(x)$ is an indicator function that equals $1$ when $x\in I_j$ and 0 otherwise, $N_{gt}$ is the number of instances in the 3D scene, $M$ is the number of $p^o$, and ${\text{cnt}}_j$ is the
center of $j$-th instance $I_j$.

\subsubsection{Detection Loss.}
%Following the implementation details of previous works~\cite{3detr, chen2023vote2capdetr}, 
Following DETR~\cite{carion2020detr}, we use Hungarian matching~\cite{kuhn1955hungarian} to assign each proposal with the ground truth. 
The detection loss $\mathcal{L}_{\text{det}}$ is defined as:
\begin{equation}
\mathcal{L}_{\text{det}}=\alpha_1\mathcal{L}_{\text{giou}}+\alpha_2\mathcal{L}_{\text{cls}}+\alpha_3\mathcal{L}_{\text{cnt}}+\alpha_4\mathcal{L}_{\text{size}},
\end{equation}
with $\alpha_1=10, \alpha_2=1, \alpha_3=5, \alpha_4=1$ set heuristically.
This loss is applied across all decoder layers for better convergence.

\subsubsection{Caption Loss.}
%Following the standard protocol for image captioning, we train our caption head with standard cross-entropy loss (MLE training) first, then fine-tune it with Self-Critical Sequence Training (SCST)~\cite{scst}.
Following the standard image captioning protocol, we first train the caption heads using cross-entropy loss for Maximum Likelihood Estimation (MLE). 
%In the MLE training, the model learns to predict the $(t+1)$-th word $c^{t+1}_n$ based on the first $t$ words $c^{[1:t]}_n$ and the visual features $\mathcal{V}$.
%The loss function is established for the final caption with length $T$ is defined as follows:
During MLE training, the model predicts the \((t+1)\)-th word \(c^{t+1}_n\) based on the first \(t\) words \(c^{[1:t]}_n\) and the visual features \(\mathcal{V}\). 
The loss function for a final caption of length \(T\) is defined as follows:
\begin{equation}
    \mathcal{L}_{c_n}=\sum_{t=1}^T\mathcal{L}_{c_n}(t)=-\sum_{t=1}^T\log{\hat{P}\bigg(c_{n}^{t+1}\vert\mathcal{V},c_{n}^{[1:t]}\bigg)}.
\end{equation}

%After the caption head is trained with word-level supervision, it is further refined using Self-Critical Sequence Training (SCST)~\cite{scst}. During this phase, the model generates multiple captions \(\hat{c}_{1,...,k}\) using a beam size of \(k\), along with an additional caption \(\hat{g}\) produced via greedy search as a baseline.
After word-level supervision, the caption head is refined using Self-Critical Sequence Training (SCST)~\cite{scst}, where the model generates multiple captions \(\hat{c}_{1,...,k}\) with a beam size of \(k\) and an additional caption \(\hat{g}\) via greedy search. 
The loss function for SCST is formulated as follows:
\begin{equation}
    \mathcal{L}_{c_n}=-\sum_{i=1}^k(R(\hat{c_i})-R(\hat{g}))\cdot\frac{1}{\vert\hat{c_i}\vert}\log{\hat{P}(\hat{c}_i\vert\mathcal{V})}.
\end{equation}
The reward function $R(\cdot)$ is based on the CIDEr~\cite{vedantam2015cider} metric for caption evaluation, and the logarithmic probability of the caption $\hat{c}_i$ is normalized by its length $|\hat{c}_i|$, ensuring equal importance to captions of different lengths.

\subsubsection{Final Loss.}
Given the instance query Loss $\mathcal{L}^o$, the detection loss for the $i$-th decoder layer as $\mathcal{L}^i_{\text{det}}$, and the average of the caption loss $\mathcal{L}_{c_n}$ within a batch, denoted as $\mathcal{L}_{\text{cap}}$, 
the final loss $\mathcal{L}$ is formulated as:
\begin{equation}
\mathcal{L}=\beta_1 \mathcal{L}^o+\beta_2 \sum_{i=1}^{n_{\text {dec-layer }}} \mathcal{L}^i_{\text {det}}+\beta_3 \mathcal{L}_{\text {cap }},
\end{equation}
where $\beta_{1}=10$, $\beta_{2}=1$, and $\beta_{3}=5$.
\section{Experiments}
\label{sec:experiments}

\begin{comment}
In this section, we first explain the datasets and evaluation metrics for 3D dense captioning in the \Cref{sec:datasets_and_metrics}.
Then, we describe implementation details for our methods in the \Cref{sec:implementation_details}.
In the \Cref{sec:comparison_with_existing_methods} and \Cref{sec:qualitative_analysis}, we conduct experiments using two benchmark datasets \cite{chen2020scanrefer, achlioptas2020referit3d} and present the results by comparing with the state-of-the-art methods.
Finally, we provide ablation studies to demonstrate the effectiveness of each component of our model and analyze the results in the \Cref{sec:ablation_study}.
\end{comment}

\subsection{Datasets and Metrics}
\label{sec:datasets_and_metrics}

\subsubsection{Datasets.}
We focus on 3D dense captioning, leveraging two benchmark datasets: ScanRefer~\cite{chen2020scanrefer} and Nr3D~\cite{achlioptas2020referit3d}. 
These datasets offer an extensive human-generated description of 3D scenes and objects.
ScanRefer encompasses $36,665$ descriptions covering $7,875$ objects within $562$ scenes, while Nr3D contains $32,919$ descriptions for $4,664$ objects across $511$ scenes. 
Both datasets draw their training data from the ScanNet~\cite{dai2017scannet} database, which includes $1,201$ 3D scenes.
For evaluation, %we utilize a subset of these datasets.
we use $9,508$ descriptions for $2,068$ objects across $141$ scenes from ScanRefer and $8,584$ descriptions for $1,214$ objects in $130$ scenes from Nr3D, all sourced from the $312$ 3D scenes in the ScanNet validation set.

\subsubsection{Metrics.}
We evaluate model performance using four metrics: CIDEr~\cite{vedantam2015cider}, BLEU-4~\cite{papineni2002bleu}, METEOR~\cite{banerjee2005meteor}, and ROUGE-L~\cite{chin2004rouge}, denoted as \textbf{C}, \textbf{B-4}, \textbf{M}, and \textbf{R}, respectively.
Following the previous studies \cite{chen2021scan2cap, cai20223djcg, jiao2022more, wang2022spacap3d, chen2023vote2capdetr}, Non-Maximum Suppression (NMS) is initially applied to filter out duplicate object predictions among the proposals. 
Each proposal is represented as a pair consisting of a predicted bounding box $\hat{b}{i}$ and its associated caption $\hat{c}{i}$. To accurately evaluate the model's capacity for object localization and caption generation, we employ the metric $m$@$k$, setting the IoU thresholds at $0.25$ and $0.5$ for our experiments, following \cite{chen2021scan2cap}:
\begin{equation}
m @ k = \frac{1}{N} \sum_{i=1}^N m\left(\hat{c}_i, C_i\right) \cdot \mathbb{I}\left\{\operatorname{IoU}\left(\hat{b}_i, b_i\right) \geq k\right\},
\end{equation}
where $N$ denotes the total number of annotated objects in the evaluation set, and $m$ stands for the captioning metrics C, B-4, M, and R.

\subsection{Implementation Details}
\label{sec:implementation_details}

Following the \cite{chen2023vote2capdetr}, our training comprises three stages.
We pre-train our network without the caption head on the ScanNet dataset~\cite{dai2017scannet} for $1,080$ epochs, using a batch size of $8$.
We minimize the loss function with an AdamW optimizer~\cite{adamw}, starting with a learning rate of $5\times10^{-4}$ that reduces to $10^{-6}$ using a cosine annealing schedule, along with a weight decay of 0.1 and gradient clipping at $0.1$ as per \cite{3detr}.
We then proceed to jointly train the model using standard cross-entropy loss for $720$ epochs on both ScanRefer~\cite{chen2020scanrefer} and Nr3D~\cite{achlioptas2020referit3d}, maintaining the detector's learning rate at $10^{-6}$ and reducing the caption head from $10^{-4}$ to $10^{-6}$ to avoid overfitting.
In the SCST\cite{scst}, we adjust the caption head using a batch size of $2$ while keeping the detector fixed over a span of $180$ epochs and maintain a constant learning rate of $10^{-6}$.
Additionally, for experiments incorporating 2D data, we use the pre-trained ENet~\cite{enet} to extract 128-dimensional multi-view features from ScanNet images as outlined in Scan2Cap~\cite{chen2021scan2cap}. 
The model has $16.9$M parameters, and the average inference time on ScanRefer~\cite{chen2020scanrefer} is $1.8$ms.
All experiments are conducted on a single Titan RTX GPU using PyTorch~\cite{pytorch}.

\begin{table*}[t]
   \centering
   \resizebox{\textwidth}{!}{%
     \begin{tabular}{c@{\hspace{-2pt}}c|cccc@{\hspace{30pt}}cccc@{\hspace{30pt}}cccc@{\hspace{30pt}}cccc}
     \toprule
     \multirow{1}{*}{} & \multicolumn{1}{c}{} & \multicolumn{8}{c|}{w/o additional 2D data} & \multicolumn{8}{c}{w/ additional 2D data} \\
     \multirow{1}{*}{Model} & \multicolumn{1}{c}{Training} & \multicolumn{4}{c}{IoU=$0.25$} & \multicolumn{4}{c|}{IoU=$0.50$} & \multicolumn{4}{c}{IoU=$0.25$} & \multicolumn{4}{c}{IoU=$0.50$} \\
     \cmidrule(lr){3-6} \cmidrule(lr){7-10} \cmidrule(lr){11-14} \cmidrule(lr){15-18}
     \multirow{1}{*}{} & \multicolumn{1}{c}{} 
     & \multicolumn{1}{c}{C$\uparrow$} & \multicolumn{1}{c}{B-4$\uparrow$} & \multicolumn{1}{c}{M$\uparrow$} & \multicolumn{1}{c}{R$\uparrow$} 
     & \multicolumn{1}{c}{C$\uparrow$} & \multicolumn{1}{c}{B-4$\uparrow$} & \multicolumn{1}{c}{M$\uparrow$} & \multicolumn{1}{c|}{R$\uparrow$} 
     & \multicolumn{1}{c}{C$\uparrow$} & \multicolumn{1}{c}{B-4$\uparrow$} & \multicolumn{1}{c}{M$\uparrow$} & \multicolumn{1}{c}{R$\uparrow$} 
     & \multicolumn{1}{c}{C$\uparrow$} & \multicolumn{1}{c}{B-4$\uparrow$} & \multicolumn{1}{c}{M$\uparrow$} & \multicolumn{1}{c}{R$\uparrow$} \\
     \midrule
     \midrule
     
     \multirow{1}{*}{Scan2Cap} & \multicolumn{1}{c}{} 
     & \multicolumn{1}{c}{53.73} & \multicolumn{1}{c}{34.25} & \multicolumn{1}{c}{26.14} & \multicolumn{1}{c}{54.95} 
     & \multicolumn{1}{c}{35.20} & \multicolumn{1}{c}{22.36} & \multicolumn{1}{c}{21.44} & \multicolumn{1}{c|}{43.57} 
     & \multicolumn{1}{c}{56.82} & \multicolumn{1}{c}{34.18} & \multicolumn{1}{c}{26.29} & \multicolumn{1}{c}{55.27}
     & \multicolumn{1}{c}{39.08} & \multicolumn{1}{c}{23.32} & \multicolumn{1}{c}{21.97} & \multicolumn{1}{c}{44.78} \\

     \multirow{1}{*}{D3Net} & \multicolumn{1}{c}{} 
     & \multicolumn{1}{c}{-} & \multicolumn{1}{c}{-} & \multicolumn{1}{c}{-} & \multicolumn{1}{c}{-} 
     & \multicolumn{1}{c}{-} & \multicolumn{1}{c}{-} & \multicolumn{1}{c}{-} & \multicolumn{1}{c|}{-} 
     & \multicolumn{1}{c}{-} & \multicolumn{1}{c}{-} & \multicolumn{1}{c}{-} & \multicolumn{1}{c}{-}
     & \multicolumn{1}{c}{46.07} & \multicolumn{1}{c}{30.29} & \multicolumn{1}{c}{24.35} & \multicolumn{1}{c}{51.67} \\

     \multirow{1}{*}{SpaCap3d} & \multicolumn{1}{c}{} 
     & \multicolumn{1}{c}{58.06} & \multicolumn{1}{c}{35.30} & \multicolumn{1}{c}{26.16} & \multicolumn{1}{c}{55.03} 
     & \multicolumn{1}{c}{42.76} & \multicolumn{1}{c}{25.38} & \multicolumn{1}{c}{22.84} & \multicolumn{1}{c|}{45.66} 
     & \multicolumn{1}{c}{63.30} & \multicolumn{1}{c}{36.46} & \multicolumn{1}{c}{26.71} & \multicolumn{1}{c}{55.71}
     & \multicolumn{1}{c}{44.02} & \multicolumn{1}{c}{25.26} & \multicolumn{1}{c}{22.33} & \multicolumn{1}{c}{45.36} \\

     \multirow{1}{*}{MORE} & \multicolumn{1}{c}{} 
     & \multicolumn{1}{c}{58.89} & \multicolumn{1}{c}{35.41} & \multicolumn{1}{c}{26.36} & \multicolumn{1}{c}{55.41} 
     & \multicolumn{1}{c}{38.98} & \multicolumn{1}{c}{23.01} & \multicolumn{1}{c}{21.65} & \multicolumn{1}{c|}{44.33} 
     & \multicolumn{1}{c}{62.91} & \multicolumn{1}{c}{36.25} & \multicolumn{1}{c}{26.75} & \multicolumn{1}{c}{56.33}
     & \multicolumn{1}{c}{40.94} & \multicolumn{1}{c}{22.93} & \multicolumn{1}{c}{21.66} & \multicolumn{1}{c}{44.42} \\

     \multirow{1}{*}{3DJCG} & \multicolumn{1}{c}{} 
     & \multicolumn{1}{c}{60.86} & \multicolumn{1}{c}{39.67} & \multicolumn{1}{c}{27.45} & \multicolumn{1}{c}{59.02} 
     & \multicolumn{1}{c}{47.68} & \multicolumn{1}{c}{31.53} & \multicolumn{1}{c}{24.28} & \multicolumn{1}{c|}{51.80} 
     & \multicolumn{1}{c}{64.70} & \multicolumn{1}{c}{40.17} & \multicolumn{1}{c}{27.66} & \multicolumn{1}{c}{59.23}
     & \multicolumn{1}{c}{49.48} & \multicolumn{1}{c}{31.03} & \multicolumn{1}{c}{24.22} & \multicolumn{1}{c}{50.80} \\

     \multirow{1}{*}{Contextual} & \multicolumn{1}{c}{} 
     & \multicolumn{1}{c}{-} & \multicolumn{1}{c}{-} & \multicolumn{1}{c}{-} & \multicolumn{1}{c}{-}
     & \multicolumn{1}{c}{42.77} & \multicolumn{1}{c}{23.60} & \multicolumn{1}{c}{22.05} & \multicolumn{1}{c|}{45.13}
     & \multicolumn{1}{c}{-} & \multicolumn{1}{c}{-} & \multicolumn{1}{c}{-} & \multicolumn{1}{c}{-}
     & \multicolumn{1}{c}{46.11} & \multicolumn{1}{c}{25.47} & \multicolumn{1}{c}{22.64} & \multicolumn{1}{c}{45.96} \\

     \multirow{1}{*}{REMAN} & \multicolumn{1}{c}{MLE} 
     & \multicolumn{1}{c}{-} & \multicolumn{1}{c}{-} & \multicolumn{1}{c}{-} & \multicolumn{1}{c}{-}
     & \multicolumn{1}{c}{-} & \multicolumn{1}{c}{-} & \multicolumn{1}{c}{-} & \multicolumn{1}{c|}{-} 
     & \multicolumn{1}{c}{62.01} & \multicolumn{1}{c}{36.37} & \multicolumn{1}{c}{27.76} & \multicolumn{1}{c}{56.25}
     & \multicolumn{1}{c}{45.00} & \multicolumn{1}{c}{26.31} & \multicolumn{1}{c}{22.67} & \multicolumn{1}{c}{46.96} \\
     
     \multirow{1}{*}{3D-VLP} & \multicolumn{1}{c}{} 
     & \multicolumn{1}{c}{64.09} & \multicolumn{1}{c}{39.84} & \multicolumn{1}{c}{27.65} & \multicolumn{1}{c}{58.78}
     & \multicolumn{1}{c}{50.02} & \multicolumn{1}{c}{31.87} & \multicolumn{1}{c}{24.53} & \multicolumn{1}{c|}{51.17} 
     & \multicolumn{1}{c}{70.73} & \multicolumn{1}{c}{41.03} & \multicolumn{1}{c}{28.14} & \multicolumn{1}{c}{59.72}
     & \multicolumn{1}{c}{54.94} & \multicolumn{1}{c}{32.31} & \multicolumn{1}{c}{24.83} & \multicolumn{1}{c}{51.51} \\

     \multirow{1}{*}{Vote2Cap-DETR} & \multicolumn{1}{c}{} 
     & \multicolumn{1}{c}{71.45} & \multicolumn{1}{c}{39.34} & \multicolumn{1}{c}{28.25} & \multicolumn{1}{c}{59.33}
     & \multicolumn{1}{c}{61.81} & \multicolumn{1}{c}{34.46} & \multicolumn{1}{c}{26.22} & \multicolumn{1}{c|}{54.40} 
     & \multicolumn{1}{c}{72.79} & \multicolumn{1}{c}{39.17} & \multicolumn{1}{c}{28.06} & \multicolumn{1}{c}{59.23}
     & \multicolumn{1}{c}{59.32} & \multicolumn{1}{c}{32.42} & \multicolumn{1}{c}{25.28} & \multicolumn{1}{c}{52.38} \\

     \multirow{1}{*}{Unit3D} & \multicolumn{1}{c}{} 
     & \multicolumn{1}{c}{-} & \multicolumn{1}{c}{-} & \multicolumn{1}{c}{-} & \multicolumn{1}{c}{-}
     & \multicolumn{1}{c}{-} & \multicolumn{1}{c}{-} & \multicolumn{1}{c}{-} & \multicolumn{1}{c|}{-}
     & \multicolumn{1}{c}{-} & \multicolumn{1}{c}{-} & \multicolumn{1}{c}{-} & \multicolumn{1}{c}{-}
     & \multicolumn{1}{c}{46.69} & \multicolumn{1}{c}{27.22} & \multicolumn{1}{c}{21.91} & \multicolumn{1}{c}{45.98} \\

     \multirow{1}{*}{Vote2Cap-DETR++} & \multicolumn{1}{c}{} 
     & \multicolumn{1}{c}{76.36} & \multicolumn{1}{c}{41.37} & \multicolumn{1}{c}{28.70} & \multicolumn{1}{c}{60.00}
     & \multicolumn{1}{c}{67.58} & \multicolumn{1}{c}{37.05} & \multicolumn{1}{c}{26.89} & \multicolumn{1}{c|}{55.64} 
     & \multicolumn{1}{c}{77.03} & \multicolumn{1}{c}{40.99} & \multicolumn{1}{c}{28.53} & \multicolumn{1}{c}{59.59}
     & \multicolumn{1}{c}{64.32} & \multicolumn{1}{c}{34.73} & \multicolumn{1}{c}{26.04} & \multicolumn{1}{c}{53.67} \\
          
     \multirow{1}{*}{\textbf{{\modelname} (Ours)}} & \multicolumn{1}{c}{} 
     & \multicolumn{1}{c}{\textbf{78.42}} & \multicolumn{1}{c}{\textbf{41.46}} & \multicolumn{1}{c}{\textbf{28.82}} & \multicolumn{1}{c}{\textbf{60.02}} 
     & \multicolumn{1}{c}{\textbf{68.46}} & \multicolumn{1}{c}{\textbf{38.23}} & \multicolumn{1}{c}{\textbf{27.56}} & \multicolumn{1}{c|}{\textbf{58.56}} 
     & \multicolumn{1}{c}{\textbf{78.35}} & \multicolumn{1}{c}{\textbf{41.20}} & \multicolumn{1}{c}{\textbf{28.82}} & \multicolumn{1}{c}{\textbf{59.80}}
     & \multicolumn{1}{c}{\textbf{66.47}} & \multicolumn{1}{c}{\textbf{36.13}} & \multicolumn{1}{c}{\textbf{26.71}} & \multicolumn{1}{c}{\textbf{54.54}} \\
     
     \midrule

     \multirow{1}{*}{Scan2Cap} & \multicolumn{1}{c}{} 
     & \multicolumn{1}{c}{-} & \multicolumn{1}{c}{-} & \multicolumn{1}{c}{-} & \multicolumn{1}{c}{-} 
     & \multicolumn{1}{c}{-} & \multicolumn{1}{c}{-} & \multicolumn{1}{c}{-} & \multicolumn{1}{c|}{-} 
     & \multicolumn{1}{c}{-} & \multicolumn{1}{c}{-} & \multicolumn{1}{c}{-} & \multicolumn{1}{c}{-}
     & \multicolumn{1}{c}{48.38} & \multicolumn{1}{c}{26.09} & \multicolumn{1}{c}{22.15} & \multicolumn{1}{c}{44.74} \\

     \multirow{1}{*}{D3Net} & \multicolumn{1}{c}{} 
     & \multicolumn{1}{c}{-} & \multicolumn{1}{c}{-} & \multicolumn{1}{c}{-} & \multicolumn{1}{c}{-} 
     & \multicolumn{1}{c}{-} & \multicolumn{1}{c}{-} & \multicolumn{1}{c}{-} & \multicolumn{1}{c|}{-} 
     & \multicolumn{1}{c}{-} & \multicolumn{1}{c}{-} & \multicolumn{1}{c}{-} & \multicolumn{1}{c}{-}
     & \multicolumn{1}{c}{62.64} & \multicolumn{1}{c}{35.68} & \multicolumn{1}{c}{25.72} & \multicolumn{1}{c}{53.90} \\

     \multirow{1}{*}{$\chi$-Tran2Cap} & \multicolumn{1}{c}{} 
     & \multicolumn{1}{c}{58.81} & \multicolumn{1}{c}{34.17} & \multicolumn{1}{c}{25.81} & \multicolumn{1}{c}{54.10} 
     & \multicolumn{1}{c}{41.52} & \multicolumn{1}{c}{23.83} & \multicolumn{1}{c}{21.90} & \multicolumn{1}{c|}{44.97} 
     & \multicolumn{1}{c}{61.83} & \multicolumn{1}{c}{35.65} & \multicolumn{1}{c}{26.61} & \multicolumn{1}{c}{54.70}
     & \multicolumn{1}{c}{43.87} & \multicolumn{1}{c}{25.05} & \multicolumn{1}{c}{22.46} & \multicolumn{1}{c}{45.28} \\

     \multirow{1}{*}{Contextual} & \multicolumn{1}{c}{SCST} 
     & \multicolumn{1}{c}{-} & \multicolumn{1}{c}{-} & \multicolumn{1}{c}{-} & \multicolumn{1}{c}{-}
     & \multicolumn{1}{c}{50.29} & \multicolumn{1}{c}{25.64} & \multicolumn{1}{c}{22.57} & \multicolumn{1}{c|}{44.71} 
     & \multicolumn{1}{c}{-} & \multicolumn{1}{c}{-} & \multicolumn{1}{c}{-} & \multicolumn{1}{c}{-}
     & \multicolumn{1}{c}{54.30} & \multicolumn{1}{c}{27.24} & \multicolumn{1}{c}{23.30} & \multicolumn{1}{c}{45.81} \\

     \multirow{1}{*}{Vote2Cap-DETR} & \multicolumn{1}{c}{} 
     & \multicolumn{1}{c}{84.15} & \multicolumn{1}{c}{42.51} & \multicolumn{1}{c}{28.47} & \multicolumn{1}{c}{59.26} 
     & \multicolumn{1}{c}{73.77} & \multicolumn{1}{c}{38.21} & \multicolumn{1}{c}{26.64} & \multicolumn{1}{c|}{54.71} 
     & \multicolumn{1}{c}{86.28} & \multicolumn{1}{c}{42.64} & \multicolumn{1}{c}{28.27} & \multicolumn{1}{c}{59.07}
     & \multicolumn{1}{c}{70.63} & \multicolumn{1}{c}{35.69} & \multicolumn{1}{c}{25.51} & \multicolumn{1}{c}{52.28} \\
     
     \multirow{1}{*}{Vote2Cap-DETR++} & \multicolumn{1}{c}{} 
     & \multicolumn{1}{c}{88.28} & \multicolumn{1}{c}{44.07} & \multicolumn{1}{c}{28.75} & \multicolumn{1}{c}{59.89}
     & \multicolumn{1}{c}{78.16} & \multicolumn{1}{c}{39.72} & \multicolumn{1}{c}{26.94} & \multicolumn{1}{c|}{55.52} 
     & \multicolumn{1}{c}{88.56} & \multicolumn{1}{c}{43.30} & \multicolumn{1}{c}{28.64} & \multicolumn{1}{c}{59.19}
     & \multicolumn{1}{c}{74.44} & \multicolumn{1}{c}{37.18} & \multicolumn{1}{c}{26.20} & \multicolumn{1}{c}{53.30} \\

     \multirow{1}{*}{\textbf{{\modelname} (Ours)}} & \multicolumn{1}{c}{} 
     & \multicolumn{1}{c}{\textbf{89.72}} & \multicolumn{1}{c}{\textbf{44.97}} & \multicolumn{1}{c}{\textbf{28.96}} & \multicolumn{1}{c}{\textbf{60.69}}
     & \multicolumn{1}{c}{\textbf{80.14}} & \multicolumn{1}{c}{\textbf{40.16}} & \multicolumn{1}{c}{\textbf{27.76}} & \multicolumn{1}{c|}{\textbf{56.10}} 
     & \multicolumn{1}{c}{\textbf{89.34}} & \multicolumn{1}{c}{\textbf{44.56}} & \multicolumn{1}{c}{\textbf{28.74}} & \multicolumn{1}{c}{\textbf{59.33}}
     & \multicolumn{1}{c}{\textbf{76.34}} & \multicolumn{1}{c}{\textbf{37.34}} & \multicolumn{1}{c}{\textbf{26.60}} & \multicolumn{1}{c}{\textbf{54.00}} \\

     \bottomrule
     \end{tabular}
     }
 \caption{Experimental results on the ScanRefer~\cite{chen2020scanrefer}. C, B-4, M, and R represent the captioning metrics CIDEr \cite{vedantam2015cider}, BLEU-4 \cite{papineni2002bleu}, METEOR \cite{banerjee2005meteor}, and ROUGE-L \cite{chin2004rouge}, respectively. A higher score for each indicates better performance.
 %C: CIDEr, B-4: BLEU-4, M: METEOR, R: ROUGE
 }
 \label{tab:scanrefer}
 \end{table*} 
\begin{table*}[t]
   \centering
   \resizebox{.5\linewidth}{!}{%
     \begin{tabular}{cc|cccc}
     \toprule
     \multirow{1}{*}{Model} & \multicolumn{1}{c}{Training} & \multicolumn{1}{c}{C@$0.5\uparrow$} & \multicolumn{1}{c}{B-4@$0.5\uparrow$} & \multicolumn{1}{c}{M@$0.5\uparrow$} & \multicolumn{1}{c}{R@$0.5\uparrow$} \\
     \midrule
     \midrule
     
     \multirow{1}{*}{Scan2Cap} & \multicolumn{1}{c}{} & \multicolumn{1}{c}{27.47} & \multicolumn{1}{c}{17.24} & \multicolumn{1}{c}{21.80} & \multicolumn{1}{c}{49.06} \\

     \multirow{1}{*}{D3Net} & \multicolumn{1}{c}{} & \multicolumn{1}{c}{33.85} & \multicolumn{1}{c}{20.70} & \multicolumn{1}{c}{23.13} & \multicolumn{1}{c}{53.38} \\

     \multirow{1}{*}{SpaCap3d} & \multicolumn{1}{c}{} & \multicolumn{1}{c}{33.71} & \multicolumn{1}{c}{19.92} & \multicolumn{1}{c}{22.61} & \multicolumn{1}{c}{50.50} \\

     \multirow{1}{*}{3DJCG} & \multicolumn{1}{c}{} & \multicolumn{1}{c}{38.06} & \multicolumn{1}{c}{22.82} & \multicolumn{1}{c}{23.77} & \multicolumn{1}{c}{52.99} \\

     \multirow{1}{*}{Contextual} & \multicolumn{1}{c}{MLE} & \multicolumn{1}{c}{35.26} & \multicolumn{1}{c}{20.42} & \multicolumn{1}{c}{22.77} & \multicolumn{1}{c}{50.78} \\

     \multirow{1}{*}{REMAN} & \multicolumn{1}{c}{} & \multicolumn{1}{c}{34.81} & \multicolumn{1}{c}{20.37} & \multicolumn{1}{c}{23.01} & \multicolumn{1}{c}{50.99} \\

     \multirow{1}{*}{Vote2Cap-DETR} & \multicolumn{1}{c}{} & \multicolumn{1}{c}{43.84} & \multicolumn{1}{c}{26.68} & \multicolumn{1}{c}{25.41} & \multicolumn{1}{c}{54.43} \\
     
     \multirow{1}{*}{Vote2Cap-DETR++} & \multicolumn{1}{c}{} & \multicolumn{1}{c}{47.08} & \multicolumn{1}{c}{27.70} & \multicolumn{1}{c}{25.44} & \multicolumn{1}{c}{55.22} \\

     \multirow{1}{*}{\textbf{{\modelname} (Ours)}} & \multicolumn{1}{c}{} & \multicolumn{1}{c}{\textbf{48.77}} & \multicolumn{1}{c}{\textbf{28.35}} & \multicolumn{1}{c}{\textbf{25.60}} & \multicolumn{1}{c}{\textbf{55.81}} \\
     
     \midrule

     \multirow{1}{*}{D3Net} & \multicolumn{1}{c}{} & \multicolumn{1}{c}{38.42} & \multicolumn{1}{c}{22.22} & \multicolumn{1}{c}{24.74} & \multicolumn{1}{c}{54.37} \\

     \multirow{1}{*}{$\chi$-Tran2Cap} & \multicolumn{1}{c}{} & \multicolumn{1}{c}{33.62} & \multicolumn{1}{c}{19.29} & \multicolumn{1}{c}{22.27} & \multicolumn{1}{c}{50.00} \\

     \multirow{1}{*}{Contextual} & \multicolumn{1}{c}{} & \multicolumn{1}{c}{37.37} & \multicolumn{1}{c}{20.96} & \multicolumn{1}{c}{22.89} & \multicolumn{1}{c}{51.11} \\
          
     \multirow{1}{*}{Vote2Cap-DETR} & \multicolumn{1}{c}{SCST} & \multicolumn{1}{c}{45.53} & \multicolumn{1}{c}{26.88} & \multicolumn{1}{c}{25.43} & \multicolumn{1}{c}{54.76} \\
     
     \multirow{1}{*}{Vote2Cap-DETR++} & \multicolumn{1}{c}{} & \multicolumn{1}{c}{47.62} & \multicolumn{1}{c}{28.41} & \multicolumn{1}{c}{25.63} & \multicolumn{1}{c}{54.77} \\

     \multirow{1}{*}{\textbf{{\modelname} (Ours)}} & \multicolumn{1}{c}{} & \multicolumn{1}{c}{\textbf{49.81}} & \multicolumn{1}{c}{\textbf{28.83}} & \multicolumn{1}{c}{\textbf{25.85}} & \multicolumn{1}{c}{\textbf{56.46}} \\
     
     \bottomrule
     \end{tabular}
     }
 \caption{Experimental results on the Nr3D~\cite{achlioptas2020referit3d} with IoU threshold at $0.5$. 
 %C, B-4, M, and R represent the captioning metrics CIDEr \cite{vedantam2015cider}, BLEU-4 \cite{papineni2002bleu}, METEOR \cite{banerjee2005meteor}, and ROUGE-L \cite{chin2004rouge}, respectively. %The IoU threshold is $0.5$.
 }
 \label{tab:nr3d}
 \end{table*} 
\begin{table*}[t]
   \centering
   \resizebox{.6\linewidth}{!}{%
     \begin{tabular}{c@{\hspace{20pt}}c@{\hspace{10pt}}c@{\hspace{20pt}}c@{\hspace{20pt}}c@{\hspace{20pt}}c@{\hspace{20pt}}c@{\hspace{20pt}}c}
     \toprule
     \multirow{1}{*}{} & \multicolumn{6}{c}{IoU=$0.50$} \\
     \cmidrule(lr){2-7}
     \multirow{1}{*}{Model} & \multicolumn{1}{c}{C$\uparrow$} & \multicolumn{1}{c}{B-4$\uparrow$} & \multicolumn{1}{c}{M$\uparrow$} & \multicolumn{1}{c}{R$\uparrow$} & \multicolumn{1}{c}{mAP$\uparrow$} & \multicolumn{1}{c}{AR$\uparrow$} \\
     \midrule
     \midrule

     \multirow{1}{*}{Vote2Cap-DETR~\cite{chen2023vote2capdetr}}
     & \multicolumn{1}{c}{73.77} & \multicolumn{1}{c}{38.21} & \multicolumn{1}{c}{26.64} & \multicolumn{1}{c}{54.71} & \multicolumn{1}{c}{45.56} & \multicolumn{1}{c}{67.77} \\

     \multirow{1}{*}{{\modelname} using only $V^o$}
     & \multicolumn{1}{c}{74.90} & \multicolumn{1}{c}{40.67} & \multicolumn{1}{c}{26.76} & \multicolumn{1}{c}{55.31} & \multicolumn{1}{c}{50.12} & \multicolumn{1}{c}{69.49} \\

     \midrule
     \multirow{1}{*}{Vote2Cap-DETR++~\cite{chen2024vote2cap}}
     & \multicolumn{1}{c}{78.16} & \multicolumn{1}{c}{39.72} & \multicolumn{1}{c}{26.94} & \multicolumn{1}{c}{55.52} & \multicolumn{1}{c}{55.48} & \multicolumn{1}{c}{70.89} \\

     \multirow{1}{*}{{\modelname} using $V^o$, KNN($V^c$)}
     & \multicolumn{1}{c}{79.03} & \multicolumn{1}{c}{41.36} & \multicolumn{1}{c}{27.31} & \multicolumn{1}{c}{56.76} & \multicolumn{1}{c}{55.95} & \multicolumn{1}{c}{69.62} \\

     \multirow{1}{*}{{\modelname} using $V^o$, $V^c_a$}
     & \multicolumn{1}{c}{81.22} & \multicolumn{1}{c}{40.90} & \multicolumn{1}{c}{27.39} & \multicolumn{1}{c}{57.95} & \multicolumn{1}{c}{56.91} & \multicolumn{1}{c}{70.38} \\

     \multirow{1}{*}{\textbf{{\modelname} using $V^o$, $V^c_a$, $V^o_a$ (Ours)}}
     & \multicolumn{1}{c}{\textbf{85.14}} & \multicolumn{1}{c}{\textbf{42.27}} & \multicolumn{1}{c}{\textbf{27.98}} & \multicolumn{1}{c}{\textbf{59.37}} & \multicolumn{1}{c}{\textbf{57.58}} & \multicolumn{1}{c}{\textbf{72.68}} \\

     %  \multirow{1}{*}{\textbf{{\modelname} using $V^o$, $V^c_a$, $V^o_a$ (Ours)}}
     % & \multicolumn{1}{c}{\textbf{83.14}} & \multicolumn{1}{c}{\textbf{42.16}} & \multicolumn{1}{c}{\textbf{27.76}} & \multicolumn{1}{c}{\textbf{57.67}}  \multicolumn{1}{c}{\textbf{52.59}} & \multicolumn{1}{c}{\textbf{70.52}} \\
     \bottomrule
     \end{tabular}
 }
 \caption{Ablation study on the ScanRefer~\cite{chen2020scanrefer}. 
 The core components of our {\modelname} are i) decomposition of the query set into Instance Query $V^o$ and Context Query $V^c$, ii) the Object-aware Context feature $V^c_a$, and iii) the Context-aware object feature $V^o_a$.
 }
 \label{tab:ablation_key_components}
 \end{table*} 

\subsection{Comparison with Existing Methods}
\label{sec:comparison_with_existing_methods}

In this section, we evaluate our performance against state-of-the-art methods: Scan2Cap~\cite{chen2021scan2cap}, D3Net~\cite{chen2022d3net}, SpaCap3D~\cite{wang2022spacap3d}, MORE~\cite{jiao2022more}, 3DJCG~\cite{cai20223djcg}, Contextual~\cite{zhong2022contextual}, REMAN~\cite{reman}, 3D-VLP~\cite{jin20233dvlp}, $\chi$-Tran2Cap~\cite{yuan2022xtrans2cap}, Vote2Cap-DETR~\cite{chen2023vote2capdetr}, Unit3D~\cite{chen2023unit3d}, and Vote2Cap-DETR++~\cite{chen2024vote2cap}. %using the metrics CIDEr~\cite{vedantam2015cider}, METEOR~\cite{banerjee2005meteor}, BLEU-4~\cite{papineni2002bleu}, and ROUGE-L~\cite{chin2004rouge}, denoted as \textbf{C}, \textbf{B-4}, \textbf{M}, and \textbf{R}, respectively.
We apply IoU thresholds of $0.25$ and $0.5$ for ScanRefer~\cite{chen2020scanrefer} as shown in \Cref{tab:scanrefer} and an IoU threshold of $0.5$ for Nr3D~\cite{achlioptas2020referit3d} indicated in \Cref{tab:nr3d}.
For the baselines, we present the evaluation results reported in the original papers, and "-" in \Cref{tab:scanrefer} and \Cref{tab:nr3d} indicates that such results have not been reported in the original paper.

\subsubsection{ScanRefer.}
%\Cref{tab:scanrefer} shows the results on the ScanRefer dataset.
Descriptions in the ScanRefer include the target object's attributes and spatial relationships with surrounding objects. %within the same space.
%As shown in \Cref{tab:scanrefer}, our method significantly outperforms the state-of-the-art methods across all input data settings and IoU threshold configurations.
%We attribute the significant improvement in performance to our approach of modeling context features, which understand geometric structures, separately from object features and integrating them over two stages.
As \Cref{tab:scanrefer} shows, our method outperforms existing methods in all data settings and IoU thresholds, thanks to our multi-stage contextual attention method.

%\vspace{-5pt}
\subsubsection{Nr3D.}
The Nr3D dataset evaluates the model's proficiency in interpreting human-spoken, free-form object descriptions. Our method shows a notable performance improvement over existing models in generating diverse object descriptions, as indicated in \Cref{tab:nr3d}. 

%The Nr3D dataset is designed to assess the model's proficiency in understanding human-spoken, free-form descriptions of objects.
%Our method quantitatively demonstrates its ability to generate diverse descriptions of objects by showing a significant performance improvement over existing models across all evaluation metrics, as indicated in \Cref{tab:nr3d}.
\subsection{Ablation Study and Discussion}
\label{sec:ablation_study}

The core components of our method are i) decomposition of the query set into Instance Query $V^o$ and Context Query $V^c$, ii) Contextual Attention of Objects for Context (O4C) ii) Contextual Attention of Contexts for Object (C4O).
%In the ablation study, 
We demonstrate that all components of {\modelname} contribute positively to the final performance, as shown in \Cref{tab:ablation_key_components}.

\subsubsection{Instance Query Generator.}
We define {\modelname} using only the Instance Query Generator (i.e., {\modelname} using only $V^o$ in \Cref{tab:ablation_key_components}) as our baseline and compare it with Vote2Cap-DETR~\cite{chen2023vote2capdetr}, an object-centric transformer encoder-decoder architecture.
The major difference between our instance query generator and Vote2Cap-DETR is how we generate the query set for instances.
Vote2Cap-DETR uses farthest point sampling (FPS) to generate queries before the query coordinates are adjusted through voting.
Therefore, if the coordinates are mistakenly focused on a specific object after voting, features will be extracted from the same object.
On the other hand, our instance query generator extracts the features from the candidate coordinates after the voting. 
This improves the number of matching candidates (e.g., for $2,068$ objects in our evaluation set, our method has $1,540$ matching proposals while Vote2Cap-DETR has $1,498$).
This enhancement boosts localization performance in terms of mean Average Precision (mAP) and Average Recall (AR), which directly contributes to improving dense captioning performance.

\subsubsection{Context Query Generator.}
To enable object features to focus on localization, we independently generate context and instance queries separately.
In the Vote2Cap-DETR++~\cite{chen2024vote2cap}, the decoupled queries are projections of the object-centric queries and still entail the limitations of the object-centric design.
In contrast, our context queries capture structural information from the entire 3D scene, effectively decoupling features from object localization, as shown in \Cref{tab:ablation_key_components}.

%In order to effectively extract features for each query, we conduct ROI (Region of Interest) analysis experiments for feature extraction in \Cref{tab:analysis_radius}, expanding the ROI radius of instance query generator and context query generator. 
%As shown in \Cref{tab:analysis_radius}, setting a new set of disentangled query set for contextual information with a sufficient margin shows substantial improvement in performance while simply expanding the ROIs for each instance rather resulted in a huge performance drop. 
%This indicates that simply expanding the ROI of individual objects can not substitute the rich information that can be captured by directly attending to the context region.

\subsubsection{O4C module and C4O module.}
As shown in \Cref{tab:ablation_key_components}, utilizing the object-aware context feature $V^c_a$ and the context-aware object feature $V^o_a$ results in performance improvements across all aspects.
Interestingly, we can see that performance improves even when surrounding context information is collected using KNN (See the results of {\modelname} using only $V^o$, KNN($V^c$)).
In the setting of {\modelname} using only $V^o$, KNN($V^c$), we set $K=16$.
Additionally, aggregating contextualized object features improves the model performance significantly.
It proves that when capturing context information for captioning, adding additional object information that matches the context is helpful.

\begin{figure*}[t]

  \adjustbox{minipage=1.3em,valign=t}{\subcaption{} \label{fig:scene_color}}
  \begin{subfigure}[t]{.28\textwidth}
  \centering
  \includegraphics[height=2.3cm,width=\textwidth,valign=t]
  {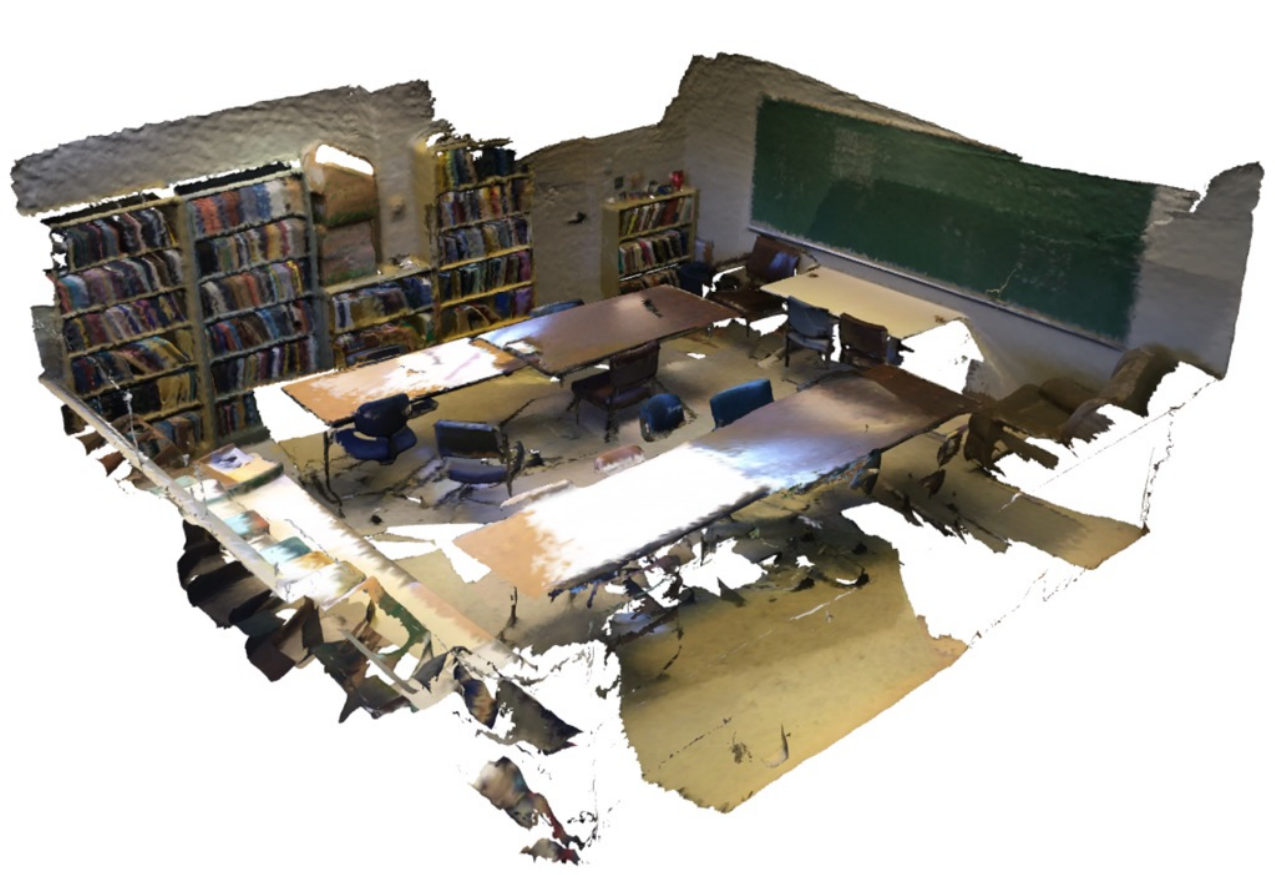}
  \end{subfigure}
  \vspace{0.25cm}
  \adjustbox{minipage=1.3em,valign=t}{\subcaption{} \label{fig:scene_point}}
  \begin{subfigure}[t]{.28\textwidth}
  \centering
  \includegraphics[height=2.3cm,width=\textwidth,valign=t]
  {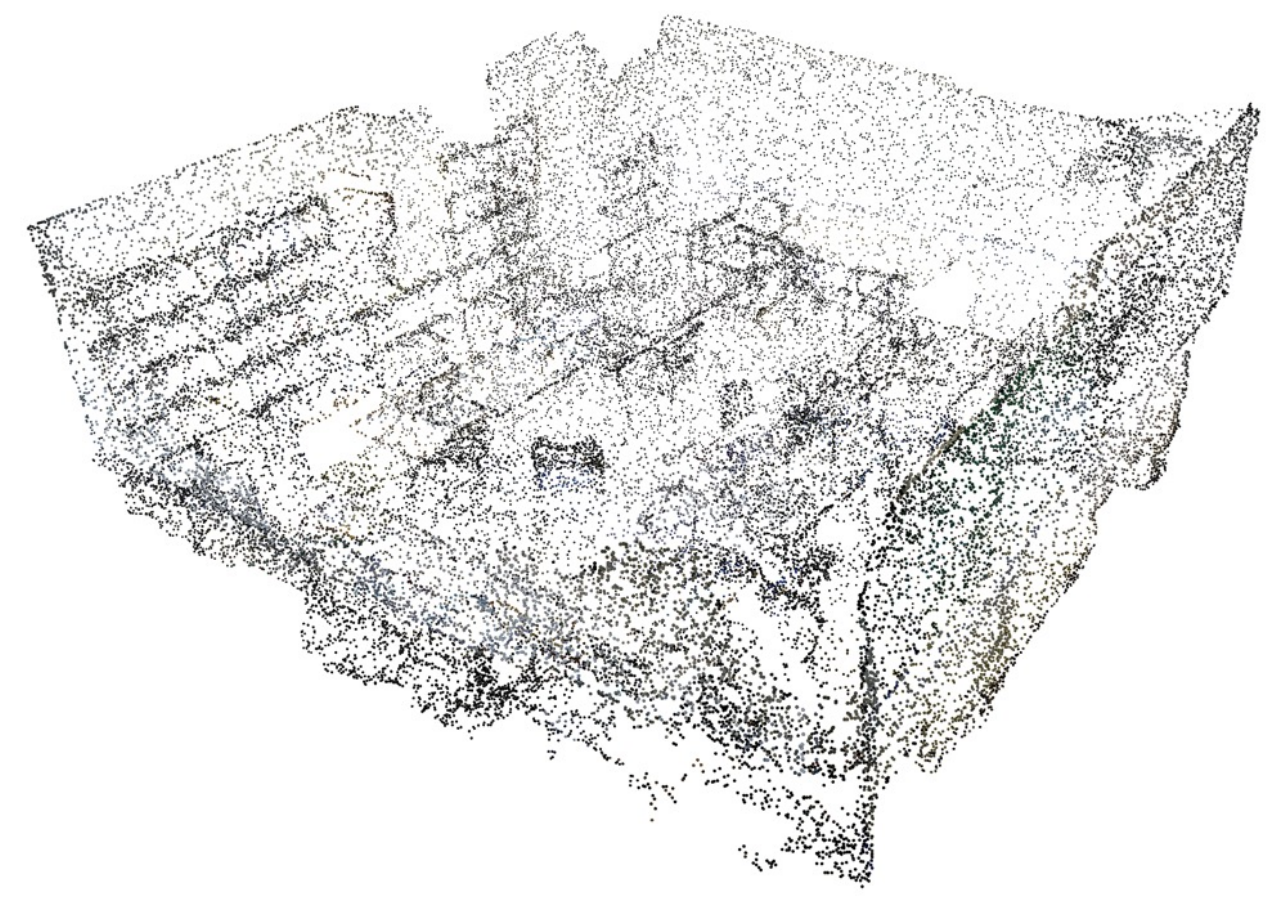}
  \end{subfigure}
  \vspace{0.25cm}
  \adjustbox{minipage=1.3em,valign=t}{\subcaption{} \label{fig:instance_query}}
  \begin{subfigure}[t]{.28\textwidth}
  \centering
  \includegraphics[height=2.3cm,width=\textwidth,valign=t]
  {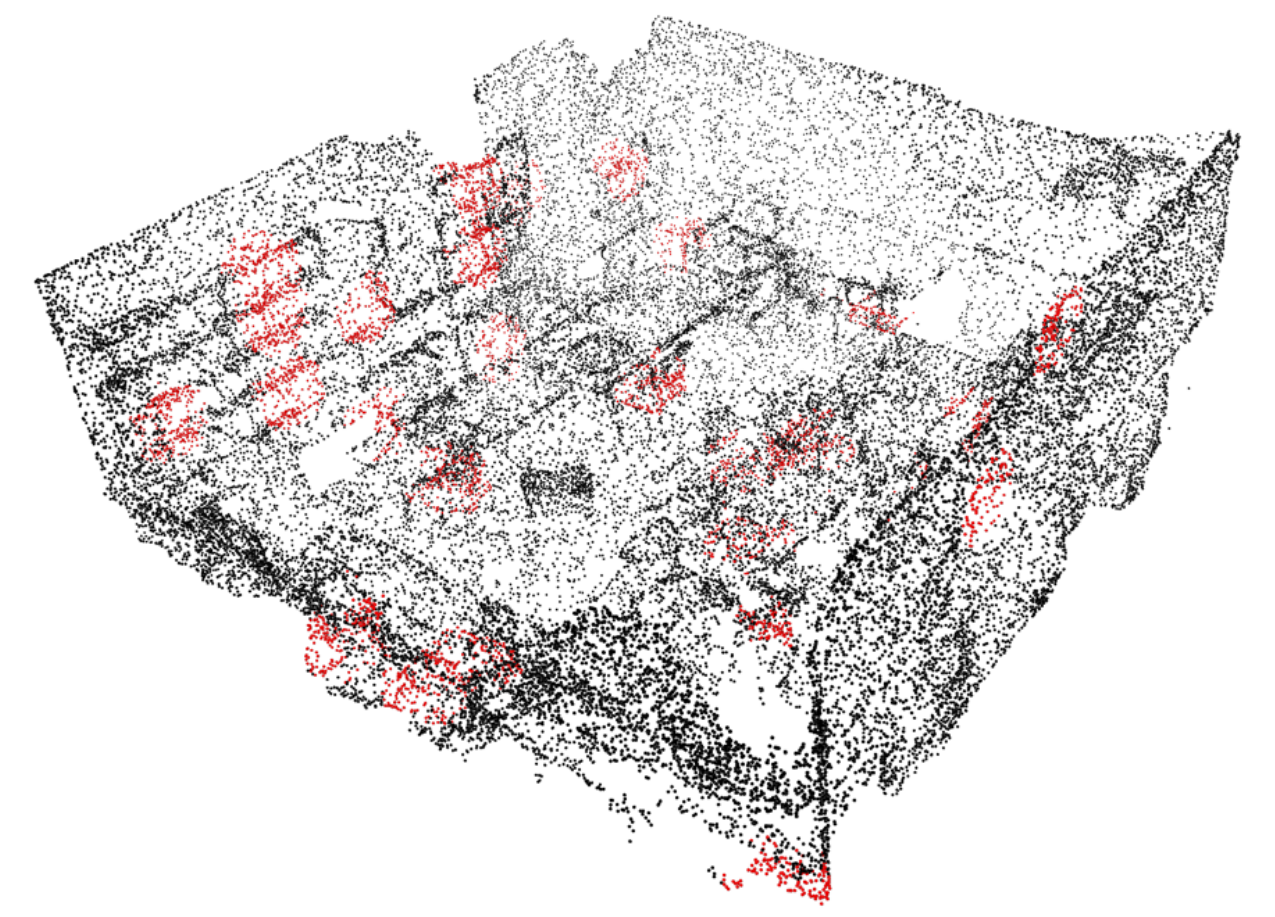}
  \end{subfigure}
  
  \adjustbox{minipage=1.3em,valign=t}{\subcaption{} \label{fig:instance_0}}
  \begin{subfigure}[t]{.28\textwidth}
  \centering
  \includegraphics[height=2.3cm,width=\textwidth,valign=t]
  {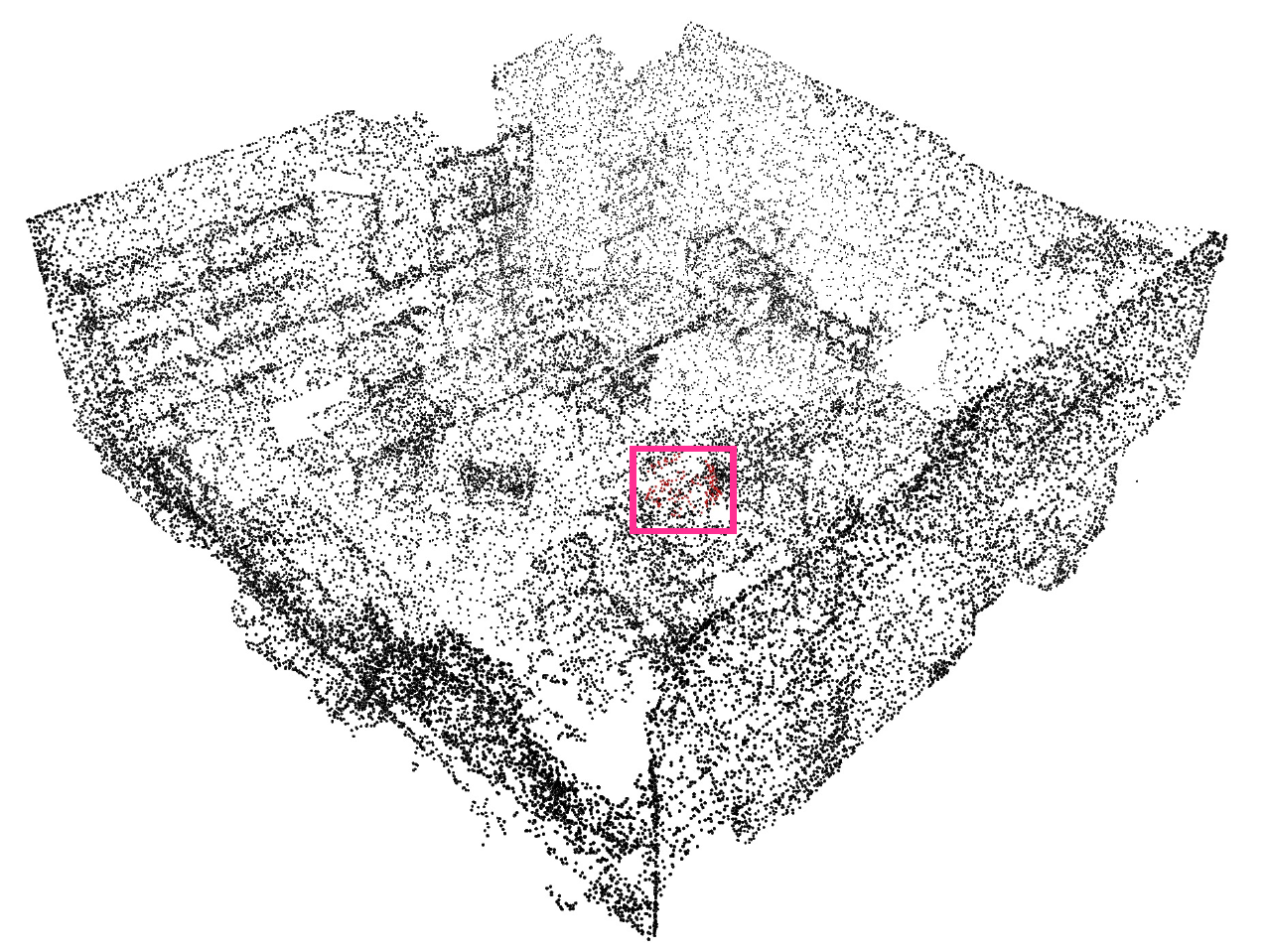}
  \end{subfigure}
  \vspace{0.25cm}
  \adjustbox{minipage=1.3em,valign=t}{\subcaption{} \label{fig:geometry_contextualization}}
  \begin{subfigure}[t]{.28\textwidth}
  \centering
  \includegraphics[height=2.3cm,width=\textwidth,valign=t]
  {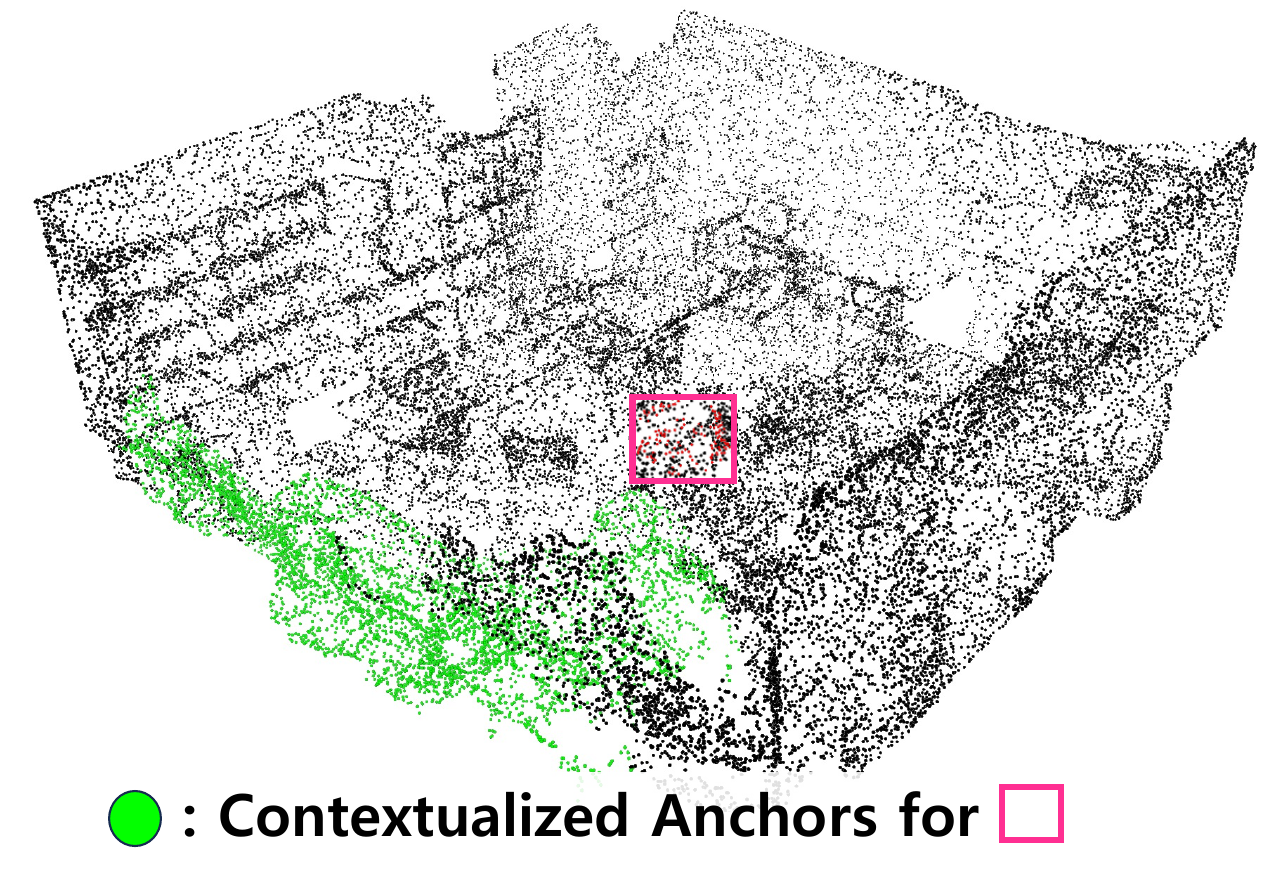}
  \end{subfigure}
  \vspace{0.25cm}
  \adjustbox{minipage=1.3em,valign=t}{\subcaption{} \label{fig:instance_contextualization}}
  \begin{subfigure}[t]{.28\textwidth}
  \centering
  \includegraphics[height=2.3cm,width=\textwidth,valign=t]
  {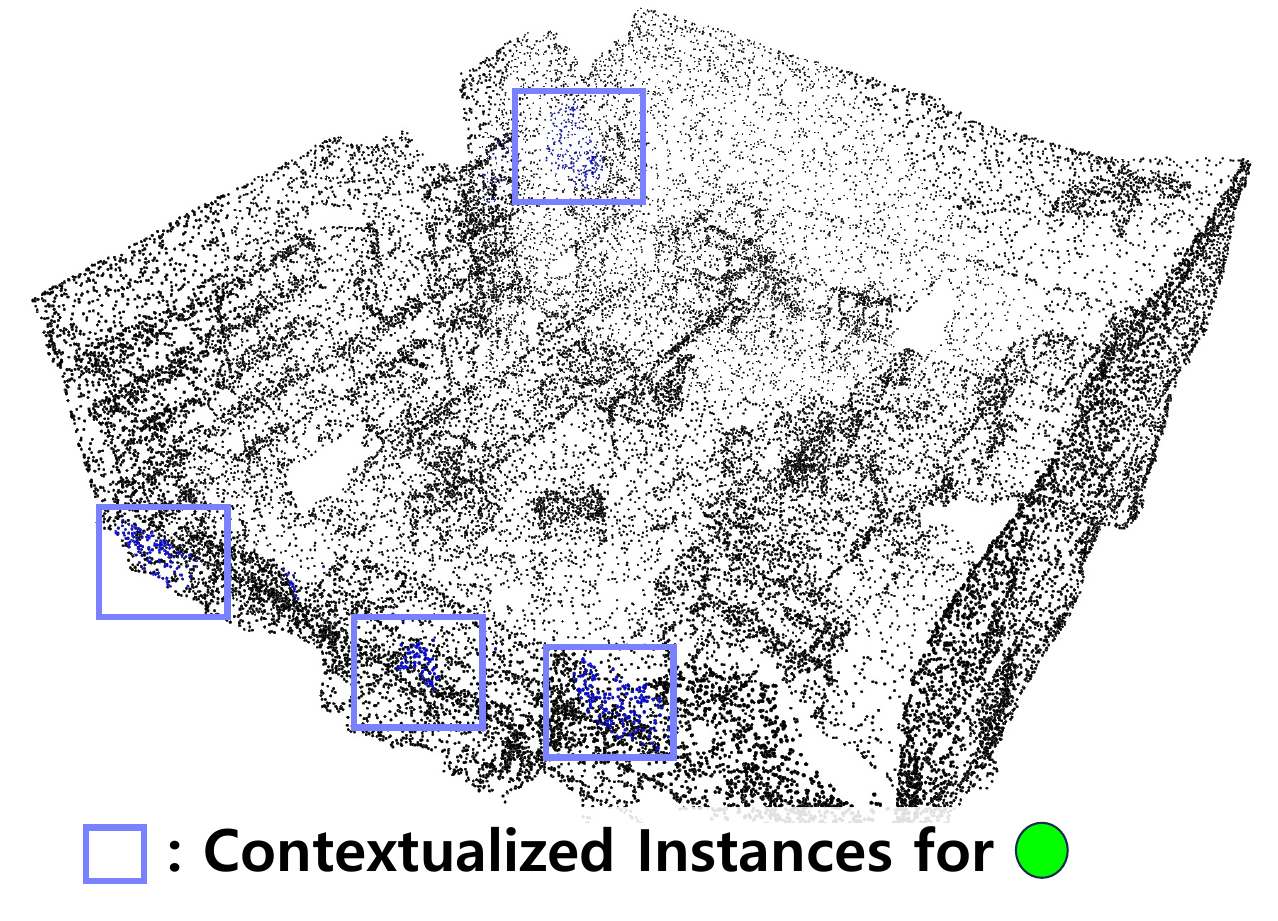}
  \end{subfigure}
  
 \caption{Visualization of (a) the input 3D scene, (b) input point cloud without color, (c) visualization of instance queries, (d) one sampled instance query, (e) geometry contextualization of (d), and (f) instance contextualization of (e) on the ScanRefer~\cite{chen2020scanrefer}.}
 \label{fig:visualization}
\end{figure*}
\begin{figure*}[t]
\begin{center}
\centerline{\includegraphics[width=0.91\textwidth]{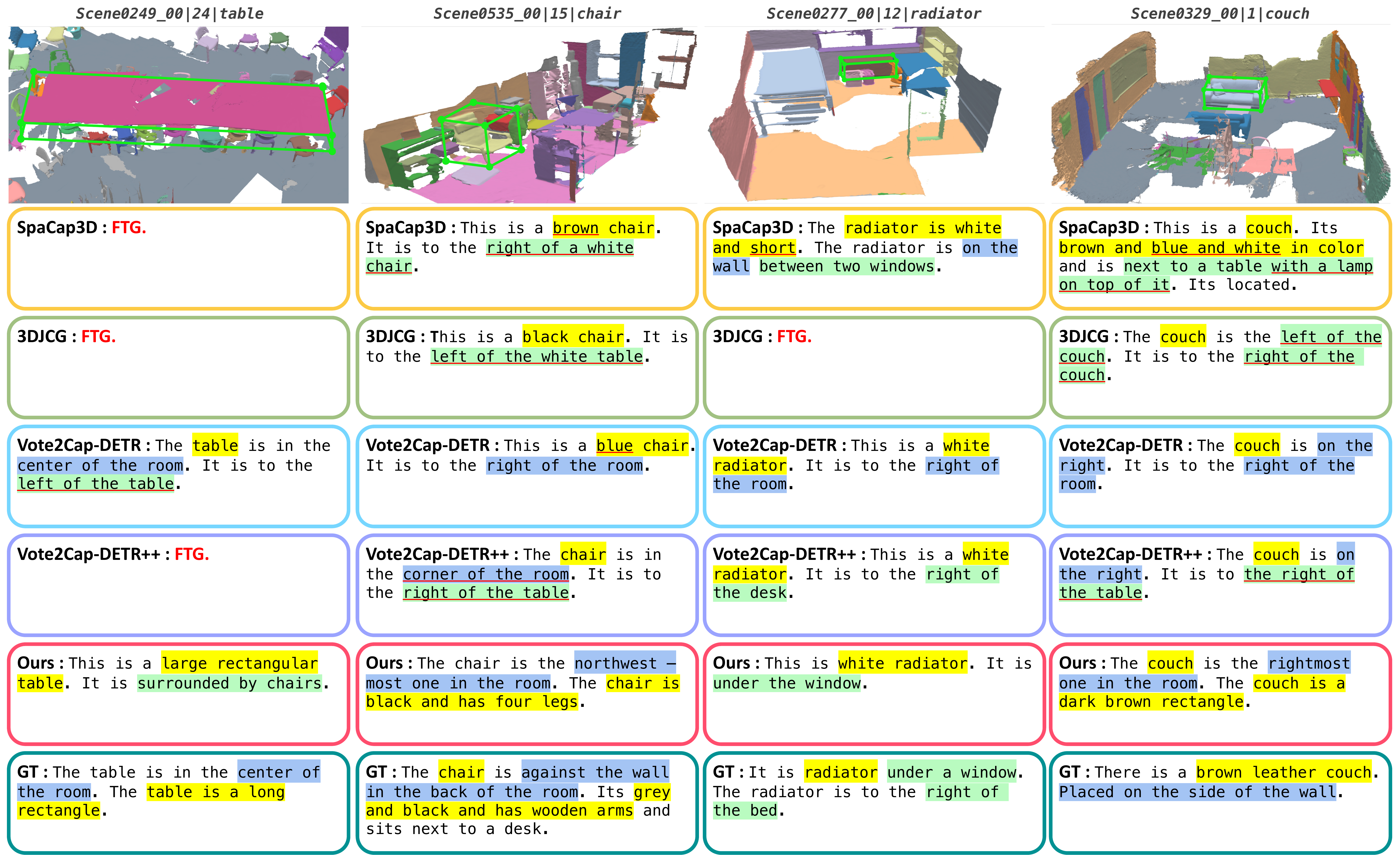}}
\caption{
Qualitative results on the ScanRefer~\cite{chen2020scanrefer}. 
The yellow-highlighted sections show information specific to the object itself, the green-highlighted sections describe the relationships between objects, and the blue-highlighted sections depict the spatial position of the object in the 3D scene. 
Captions underlined in red indicate incorrect descriptions. 
\textbf{FTG.} represent failures in caption generation due to low IoU.
}
\label{fig:qualitative_results}
\end{center}
\end{figure*}

\subsection{Qualitative Analysis}
\label{sec:qualitative_analysis}

\Cref{fig:visualization} shows the parts of the entire scene being focused on at each stage of the proposed model.
In \Cref{fig:instance_query} and \Cref{fig:instance_0}, our instance queries are not concentrated on a specific object but are spread throughout the 3D scene and exist in the location where the object exists.
\Cref{fig:geometry_contextualization} shows the part of the context queries with the highest attention score.
The green part of the \Cref{fig:geometry_contextualization} indicates the surroundings of the object and the corners of the room further away, but it needs to be related to other instances or scenes to express the situation more clearly.
The blue part of \Cref{fig:instance_contextualization} describes the part of related instance queries for the object-aware context feature selected in \Cref{fig:geometry_contextualization}.
Most instances are inside the green contextualized region with high scores of \Cref{fig:geometry_contextualization}, and some instances on the other side are also included.
This observation shows that our method can check all instances of the entire scene to generate captions for one object.

To demonstrate the effectiveness of the proposed method, we provide qualitative results with state-of-the-art models. %: SpaCap3D~\cite{wang2022spacap3d}, 3DJCG~\cite{cai20223djcg}, and Vote2Cap-DETR~\cite{chen2023vote2capdetr}.
%As shown in \Cref{fig:qualitative_results}, when a 3D scene is given as input, the model generates captions that describe objects in the given scene and evaluates the performance by comparing it with the ground-truth for each object.
%The ground-truth includes descriptions of the intrinsic properties of the object (e.g., the monitor is black and square.), explanations using the relationships between objects (e.g., a picture is hanging on the back wall with a couch below it.), and descriptions of the object in the context of the entire space (e.g., at the far end of the room.).
SpaCap3D~\cite{wang2022spacap3d}, 3DJCG~\cite{cai20223djcg}, Vote2Cap-DETR~\cite{chen2023vote2capdetr}, and Vote2Cap-DETR++~\cite{chen2024vote2cap} have attempted to incorporate contextual information by modeling relationships with object pairs or aggregating nearest neighbor features. %of an object.
In the \Cref{fig:qualitative_results}, these methods generate captions limited to the object and its immediate relations in a fixed format (e.g., this is a white radiator. it is under right of the desk).
Since their context is not collected based on the entire scene, it is challenging to localize relationships with surrounding objects or locations in the scene.
{\modelname} accurately predicts the attributes (i.e., can obtain precise representations for regions that have localization boundaries) and captions that require a sufficient understanding of structural information throughout the scene (i.e., can retrieve geometrically substantial information within the scene that exceeds the boundary of localization constraints) in various formats (e.g., the chair is the northwest – most one in the room.). %The chair is black and has four legs.).

\section{Conclusion}
\label{sec:conclusion}

In this work, we propose {\modelname}, a novel end-to-end transformer encoder-decoder pipeline with multi-stage contextual attention for 3D dense captioning.
Our {\modelname} parallelly decodes a fixed set of object queries that contain local features of individual objects and context queries which contain the non-object contexts in the 3D scene.
%For each object, we obtain a contextualized geometry feature and contextualized object feature.
%The contextualized geometry feature retrieves relevant positional features throughout the global scene, then the objects relevant to the contextualized geometry are aggregated to consist the contextualized object feature.
This allows our model to capture relevant object-aware context and context-aware object features across the entire scene without being restricted to single object localization or their immediate surroundings.
As the representation for localization and caption generation is disentangled, {\modelname} can improve both localization and contextual dense captioning performance.
We validate the effectiveness of {\modelname} by showing that it outperforms the state-of-the-art across all metrics on two benchmarks for 3D dense captioning.
%We validate the effectiveness of our method on two widely used benchmarks for 3D dense captioning: ScanRefer and Nr3D datasets, where we consistently surpass state-of-the-art performance across all metrics and datasets.

\section*{Acknowledgement}
%We sincerely thank Junhyug Noh, Seokhee Hong, Jaewoo Ahn, and Dayoon Ko for their constructive comments.
This work was supported by LG AI Research and Institute of Information \& Communications Technology Planning \& Evaluation (IITP) grant (No.~RS-2019-II191082, No.~RS-2022-II220156) funded by the Korea government (MSIT).

\bibliographystyle{splncs04}
\bibliography{main}
\end{document}